\documentclass{osa-article}

\journal{osajournal}

\articletype{Research Article}

\usepackage{graphicx}
\usepackage{cancel,xcolor}
\usepackage{hyperref}

\DeclareMathOperator*{\argmax}{arg\,max}

\newcommand{\etal}{\textit{et al}.}
\newcommand{\Cancel}[2][black]{{\color{#1}\cancel{\color{black}#2}}}

\begin{document}

\title{Invariant Descriptors for Intrinsic Reflectance Optimization}

\author{Anil S. Baslamisli\authormark{1,*} and Theo Gevers\authormark{1,2}}

\address{\authormark{1}University of Amsterdam, Science Park 904, 1098XH Amsterdam, the Netherlands\\
\authormark{2}3DUniversum, Science Park 400, 1098XH Amsterdam, the Netherlands}

\email{\authormark{*}a.s.baslamisli@uva.nl} 



\begin{abstract}
Intrinsic image decomposition aims to factorize an image into albedo (reflectance) and shading (illumination) sub-components. Being ill-posed and under-constrained, it is a very challenging computer vision problem. There are infinite pairs of reflectance and shading images that can reconstruct the same input. To address the problem, Intrinsic Images in the Wild by Bell~\etal~provides an optimization framework based on a dense conditional random field (CRF) formulation that considers long-range material relations. We improve upon their model by introducing illumination invariant image descriptors: color ratios. The color ratios and the reflectance intrinsic are both invariant to illumination and thus are highly correlated. Through detailed experiments, we provide ways to inject the color ratios into the dense CRF optimization. Our approach is physics-based, learning-free and leads to more accurate and robust reflectance decompositions.
\end{abstract}

\section{Introduction}
Intrinsic image decomposition aims to factorize an image into albedo (reflectance) and shading (illumination) sub-components~\cite{Barrow1978}. The reflectance component represents pure material colors of a scene, independent of any illumination effect including geometry and camera viewpoint. The shading component is invariant to color (albedo) and encodes the illumination effects within a scene such as shadows, shading due to geometry and ambient light. Due to their invariant properties, intrinsic images have been favored by numerous computer vision and computational photography applications. For example, semantic segmentation algorithms may profit from the illumination invariant properties of the reflectance map for robust estimations~\cite{Baslamisli2018ECCV}. In addition, intrinsic images can be used for physically plausible photo editing tasks such as recoloring~\cite{Xu2019}, color transfer and photo fusion~\cite{Beigpour2011}.

In an ideal diffuse environment, element-wise multiplication of the reflectance \emph{R} and shading \emph{S} components properly approximates the observed image \emph{I}:
\begin{equation} \label{eq:intrinsic_image}
I = R \times S\;.
\end{equation}
Being ill-posed and under-constrained, intrinsic image decomposition is a very challenging computer vision problem. Given an observed image, there exists twice as many unknowns, reflectance and shading, to predict. Hence, there are infinite pairs of reflectance and shading images that can reconstruct the same input. To address the problem, early methods adapt heuristics-based priors. For example, Land and McCann's pioneering Retinex algorithm is based on the assumption that the reflectance images can directly be estimated by strong image gradients using a surface reconstruction algorithm~\cite{Land1971}. Recent approaches either use an energy minimization (optimization) framework to constrain intrinsic image estimations based on hand-crafted priors~\cite{Bell2014,Cheng2019}, or use data driven deep learning based  techniques~\cite{Narihia2015,Baslamisli2018CVPR}.

One of the key advantages of the optimization based methods is that they do not require any labeled data, whereas deep learning based methods demand considerable amounts of training data to learn the model parameters. Optimization based methods aim to regularize the intrinsic component computation to constrain their behavior based on hand-crafted priors. Those prior are mostly based on observations. For example, global reflectance sparsity~\cite{Gehler2011,Shen2011} and piece-wise constant reflectance~\cite{Land1971,Barron2015}. However, physics-based invariant descriptors are mostly ignored for crafting priors. One example is the photometric invariant color ratios~\cite{Finlayson1992}. They eliminate shading information and thus represent an image featuring only albedo edges. In that sense, color ratios and the reflectance intrinsic are both invariant to illumination and thus are highly correlated. However, this correlation has never been explored before. Thus, to our knowledge, we are the first work to compute dense reflectance intrinsic maps using color ratios.

Therefore, in this paper, we explore the use of photometric invariant descriptors into an optimization scheme to compute reflectance (albedo) images. We demonstrate the effectiveness of such an approach by integrating color ratios into the dense CRF optimization framework of Bell~\etal~\cite{Bell2014}. Their model is publicly available~\footnote{\url{https://github.com/seanbell/intrinsic}},  and their conditional random field framework considers long-range relative material relations as in the case of color ratios. Our extensions to their code is straightforward, fully unsupervised, do not demand any additional data like depth or image sequences, and do not require any domain calibration. 

In particular, the \textbf{main contribution} of this work is not to chase the highest performance numbers, but to put forward a novel hypothesis that physics-based color ratios can be used to improve and robustify reflectance predictions. To sustain our hypothesis, (1) the ratios are integrated in the Color Retinex paradigm to provide complementary color cues. The results are particularly important as almost all the optimization-based intrinsic image decomposition methods consider the Color Retinex approach in their framework. Thus, a more robust prior can be achieved. (2) The ratios are used to define an adaptive clustering initialization mechanism. Current work using the global reflectance sparsity prior treats the number of clusters as a hyper-parameter and decides it through a grid search. However, the process is costly, not effective and dataset dependent. (3) The ratios are embedded into a pairwise term that uses hand-crafted priors based on observations. Unlike hand-crafted priors, the ratios are physics-based and do not need domain calibration. (4) Finally, we demonstrate that these additions lead to reflectance predictions that are robust to outdoor natural shadow handling. Experiments on four different datasets show the merits of the proposed method.

\section{Related Work}
\begin{figure}[h]
\includegraphics[width=1\linewidth]{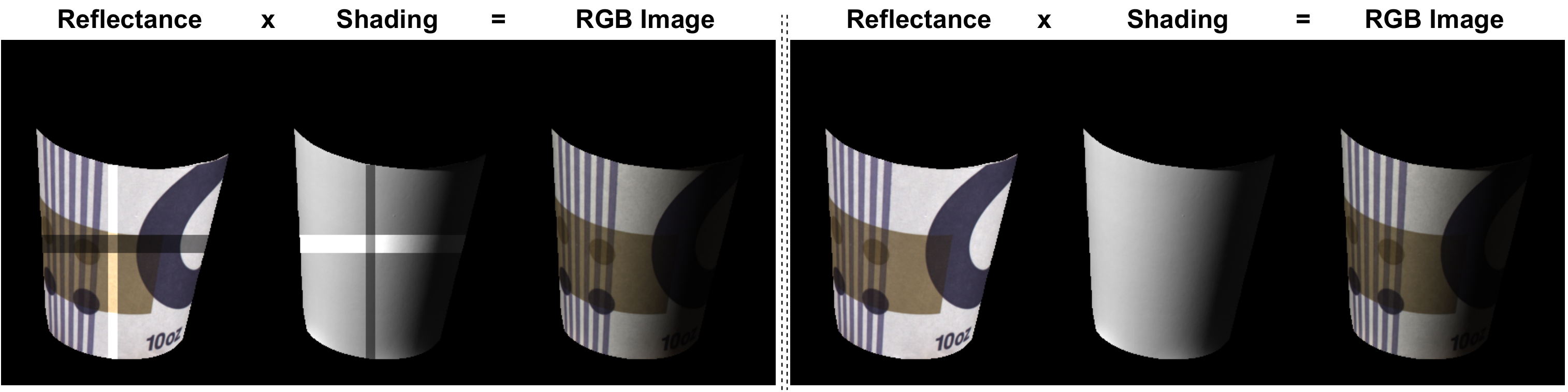}
   \caption{The ill-posed nature of the problem. The left part is an incorrect intrinsic image decomposition, whereas the right part presents the ground-truth one. Both achieve the same input image through R $\times$ S = I.}
\label{fig:illpose}
\end{figure}

The intrinsic image decomposition task is a very challenging computer vision problem as it ill-posed and under-constrained. Given a single pixel value, it seeks to find the two unknown disentangled variables; reflectance and shading. As a result, different combinations of the reflectance and shading may achieve the same input. An example is illustrated in Figure~\ref{fig:illpose}.

The pioneering work, Retinex by Land and McCann, based on the Mondrian patterns, which are composed of piece-wise distinctly colored patches, identifies strong image gradients assuming they correspond to true albedo changes~\cite{Land1971}. Then, those selected gradients are re-integrated by a surface reconstruction algorithm such as Poisson. Successive research focuses on deriving priors, based on observations, that can explain the characteristics of the intrinsic images. Usually, an optimization approach is taken to impose these constraints. For example, Gehler~\etal~and Shen~\etal~consider that the number of colors in a scene is limited, thus they sample reflectances from a sparse set of colors~\cite{Gehler2011,Shen2011}. Shen~\etal~and Zhao~\etal~observe that the the distinct points with the intensity-normalized texture structures tend to have the same reflectance values~\cite{Shen2008,Zhao2012}. In a similar manner, Shen~\etal~assume that a local neighborhood of pixels with similar intensity values also have similar reflectance values~\cite{Shen2013}. Similarly, Garces~\etal~assume that changes in chromaticity usually correspond to changes in reflectance~\cite{Garces2012}. Jiang~\etal~make the observation that correlations between local mean luminance and local luminance in textured regions indicate illumination changes~\cite{Jiang2010}. Inspired by the Retinex algorithm's assumption of piece-wise constant reflectances, other works impose piece-wise reflectance prior utilizing various regularization techniques~\cite{Barron2015,Tappen2005,Ding2017,Li2014,Sheng2020}. Additional user guidance is also explored as a prior to associate reflectance values~\cite{Shen2013,Bousseau2009}. Moreover, when available, extra sensory information can be incorporated to obtain specific priors. For instance, Chen and Koltun, Jeon~\etal~and Lee~\etal~explore supplementary depth and surface normal cues to constrain shading estimations~\cite{Chen2013,Jeon2016,Lee2012}. More recently, near-infrared imagery based priors are explored by Chenge~\etal~\cite{Cheng2019}. Finally, image sequences are utilized to impose constant reflectance prior over time~\cite{Weiss2001,Gong2019,Laffont2015}. To conclude, single image intrinsic image decomposition based on optimizations usually derive hand-crafted priors based on observation. In contrast, we propose to incorporate physics-based invariant descriptors that are not limited by the imaging conditions or the generalization quality of the hand-crafted features.

Invariant descriptors are independent on a set of imaging conditions; thus, they represent simplified versions of the observed images. To derive them, physics-based reflection models are used. An example is the color ratios by Finlayson~\cite{Finlayson1992}. As illuminant invariants, they are used to construct histogram features for object recognition. Matas~\etal~integrate ratios into a graph framework for efficient object recognition~\cite{Matas1995}. Nayar and Bolle employ color ratios for robust pose estimation~\cite{Nayar1996}. Barnard and Finlayson identify possible shadow segments by color ratios for color constancy and dynamic range compression~\cite{Barnard2000}. Color ratio gradients are also employed for content-based image retrieval of non-uniform texture objects~\cite{Gevers2001}. Later, Gevers and Smeulders extend the idea of color ratios to cross color ratios that are also invariant to the scene geometry~\cite{Gevers1997}. Not only cross color ratios are independent of illumination and geometry, but also the reflectance intrinsic by definition. Thus, cross color ratios emphasizing true color variations are expected to provide cues for intrinsic images. Nonetheless, this natural correlation between intrinsic images and cross color ratios has never been incorporated as a prior knowledge before. To that end, we suggest the use of photometric invariant descriptors into optimization to compute reflectance intrinsics. Unlike hand-crafted priors that are based on observation and require domain expertise, cross color ratios are based on physics-based reflections models. They are directly computed from the observed $RGB$ image in learning-free manner (unsupervised) and do not require additional sensory information or calibration.

Powerful deep learning based methods are also widely used for the task. Narihia~\etal~introduce the first work to predict reflectance and shading from a single $RGB$ image in an end-to-end manner~\cite{Narihia2015}. Shi~\etal~predict diffuse albedo, shading, and specular highlights from a single $RGB$ image using millions of synthetic images of objects~\cite{Shi2017}. Baslamisli~\etal~apply Retinex approach in a deep learning framework~\cite{Baslamisli2018CVPR}. Li and Snavely utilize image sequences and constant reflectance prior in a deep learning model~\cite{Li2018CVPR}. Lettry~\etal~investigate adversarial learning~\cite{Lettry2018}. Baslamisli~\etal~propose fine-grained shading decomposition~\cite{Baslamisli2019}. We refer readers to the work of Sial~\etal~that provides a comprehensive overview of the deep learning based methods and large-scale datasets~\cite{Sial2020}. 

\section{Approach}
\subsection{Image Formation Model}
To describe an image \emph{I} over the visible spectrum $\omega$, we use the diffuse dichromatic reflection model for three color channels $c \in \{R,G,B\}$ as follows~\cite{Shafer1985}:
\begin{equation} \label{eq:imf}
\begin{aligned}
I_{c} = m(\vec{n}, \vec{l}) \int_{\omega}^{} f_{c}(\lambda)\; e(\lambda)\; s(\lambda)\; \mathrm{d}\lambda \;.
\end{aligned}
\end{equation}
In the equation, \emph{m} is a function modelling the interaction between the surface normal $\vec{n}$ and the incoming light source direction $\vec{l}$ (e.g. Lambertian $m(\vec{n}, \vec{l})$ = ($\vec{n} \cdot \vec{l}$)). $\lambda$ represents the wavelength, $f$ indicates the camera spectral sensitivity, $e$ describes the spectral power distribution of the light source, and $s$ denotes the surface reflectance i.e. the albedo. Then, assuming a linear sensor response and narrow band filters ($f_{c}(\lambda_c)$)~\cite{Finlayson1994}, the equation can be simplified as follows:
\begin{equation} \label{eq:imf_mult}
I_{c} = m(\vec{n}, \vec{l}) \; e(\lambda_{c})\; s(\lambda_{c})\;=
m(\vec{n}, \vec{l}) \; e_c\; s_c\;.
\end{equation}
The equation models an image $I$ by the multiplication of surface geometry and light interaction $m(\vec{n}, \vec{l})$, reflectance $s_c$ and light source properties $e_c$ per pixel. Using Equation~\eqref{eq:imf_mult} intrinsic images are defined (per pixel) as follows:
\begin{equation} \label{eq:iid}
\begin{aligned}
&I_{c} = S_{c} \times R_{c} \;,\\
&S_{c} = m(\vec{n}, \vec{l}) \; e_{c}\;,\\
&R_{c} = s_{c}\;,
\end{aligned}
\end{equation}
\noindent where an image $I_{c}$ is modelled by the element-wise multiplication of its shading $S_{c}$ and reflectance $R_{c}$ components (intrinsics). Most works assume white light, but if the light source $e$ is colored, then that color information is embedded in the shading (illumination) component.

\subsection{Color Ratios}
For two neighboring (adjacent) pixels $x_1$ and $x_2$, locally constant illumination can be assumed $e_{c}^{x_1} = e_{c}^{x_2}$~\cite{Land1971}. Based on this property, color ratios $F$ are computed for $c \in \{R,G,B\}$ at two neighbouring pixels $x_1$ and $x_2$ as follows:
\begin{equation} \label{eq:F}
F_1 = \frac{R^{x_1}}{R^{x_2}}\;,\;\; F_2 = \frac{G^{x_1}}{G^{x_2}}\;,\;\; F_3 = \frac{B^{x_1}}{B^{x_2}}\;.
\end{equation}
We illustrate the invariant properties of (single) ratios by plugging Equation~\eqref{eq:imf_mult} into Equation~\eqref{eq:F} for $F_1$ as follows (holds also for $F_2$ and $F_3$):
\begin{equation} \label{eq:F_full}
F_1 = \frac{(m(\vec{n}, \vec{l}))^{x_1} \; \Cancel[orange]{{e_R}^{x_1}}\; {s_R}^{x_1}}{(m(\vec{n}, \vec{l}))^{x_2} \; \Cancel[orange]{{e_R}^{x_2}}\; {s_R}^{x_2}}\;.
\end{equation}
Since two neighboring pixels $x_1$ and $x_2$ share the same locally constant illumination, ${e_R}^{x_1}$ and ${e_R}^{x_2}$ are cancelled out in the equation. If we assume that neighbouring pixels share the same geometry $(m(\vec{n}, \vec{l}))^{x_1} = (m(\vec{n}, \vec{l}))^{x_2}$, i.e. locally smooth surfaces, then the ratios are invariant to photometric effects. 

\subsection{Cross Color Ratios}
To overcome the geometry constraint of Equation~\eqref{eq:F_full}, cross color ratios are preferred, which are not only independent of the illumination conditions but also invariant to the object geometry. For neighbouring pixels $x_1$ and $x_2$, cross color ratios are defined as:
\begin{equation} \label{eq:M}
\begin{aligned}
M_1 = \;\frac{R^{x_1}\;G^{x_2}}{R^{x_2}\;G^{x_1}}\;,\;\;
M_2 = \;\frac{R^{x_1}\;B^{x_2}}{R^{x_2}\;B^{x_1}}\;,\;\;
M_3 = \;\frac{G^{x_1}\;B^{x_2}}{G^{x_2}\;B^{x_1}}\;.
\end{aligned}
\end{equation}
As a result, the assumption that neighbouring points share the same geometry, i.e. locally smooth surfaces, is no longer required for the cross color ratio that is invariant to photometric effects. We illustrate the invariant properties of the cross ratios by plugging Equation~\eqref{eq:imf_mult} into Equation~\eqref{eq:M} for $M_1$ as follows (holds also for $M_2$ and $M_3$):
\begin{equation} \label{eq:M_full}
\begin{aligned}
M_1 &= \frac{\Cancel[blue]{(m(\vec{n}, \vec{l}))^{x_1}} \; \Cancel[green]{{e_R}^{x_1}}\; {s_R}^{x_1}\;\Cancel[red]{(m(\vec{n}, \vec{l}))^{x_2}} \; \Cancel[olive]{{e_G}^{x_2}}\; {s_G}^{x_2}}{\Cancel[red]{(m(\vec{n}, \vec{l}))^{x_2}} \; \Cancel[green]{{e_R}^{x_2}}\; {s_R}^{x_2}\;\Cancel[blue]{(m(\vec{n}, \vec{l}))^{x_1}} \; \Cancel[olive]{{e_G}^{x_1}}\; {s_G}^{x_1}}\;=\; \frac{{s_R}^{x_1}\;{s_G}^{x_2}}{{s_R}^{x_2}\;{s_G}^{x_1}}\;. \\
\end{aligned}
\end{equation}
Since two neighboring pixels $x_1$ and $x_2$ share the same locally constant illumination, ${e_R}^{x_1}$ and ${e_R}^{x_2}$, and ${e_G}^{x_1}$ and ${e_G}^{x_2}$ are cancelled out. Moreover, since $R^{x_1}$ and $G^{x_1}$ and $R^{x_2}$ and $G^{x_2}$ share the same local geometry, $(m(\vec{n}, \vec{l}))$ components are also cancelled out. Therefore, Equation~\eqref{eq:M} can be used to compute a measure for photometric invariance without any additional constraint. The ratio encodes relative reflectance relation between two pixels. If there is no intrinsic color change (reflectance between $x_1$ and $x_2$), then the ratio is 1. Sensor artifacts or noise may slightly deviate the value from 1. On the other hand, when the ratio deviates significantly from 1, it manifests an intrinsic color change. A threshold can be defined or learned to differentiate significant deviations; identifying true color changes. This property can also be used to measure the number of distinct color edges in an image, which is explained in Section~\ref{sec:cluster}.

\subsection{Intrinsic Image Optimization}
An overview is given of the optimization framework of~\cite{Bell2014}. The problem is formulated by finding the decomposition (reflectance $R^*$ and shading $S^*$) that most likely matches the $RGB$ image $I$ based on a number of priors under probability distribution $p$ as follows:
\begin{equation} 
\label{eq:iiw}
R^*\;,\;S^* = \argmax_{R\;,\;S} p(R\;,\;S\;|\;I)\;.
\end{equation}
A number of priors are utilized such that the framework assigns a high probability to the decompositions that are consistent with these priors. In particular, the priors are $(i)$ pixels that are nearby and have similar chromaticity or intensity should also have similar reflectance, $(ii)$ reflectances are piecewise-constant (Retinex), $(iii)$ reflectances are sampled from a sparse set (global reflectance sparsity), $(iv)$ certain shading values are a priori more likely than others, $(v)$ neighboring pixels have similar shading (shading smoothness), and $(vi)$ shading is grayscale, or the same color as the light source (color constancy). Then, the framework incorporates the priors in a global sense using a dense conditional random field (CRF).

First, a set of possible labels (colors) are selected from the observed $RGB$ image as the initial reflectance predictions using clustering. The clustering is based on the chromaticity computed from the $RGB$ image, which separates reflectances under ideal conditions. Then, each reflectance pixel is labeled with a chromaticity value chosen from the clusters such
that $p(R,S|\;I)$ is maximized. Secondly, the reflectances are tuned by minimizing the discontinuities in the shading estimations. That is achieved by the optimization process aiming at to minimize the energy function $E(x)$, composed of pairwise $E_{p}(x)$ and unary ($E_{s}(x)$ and $E_{l}(x)$) costs, by imposing the prior constraints:
\begin{equation} \label{eq:energy_function}
E(x) = \omega_p\;E_{p}(x) + \omega_s\;E_{s}(x) + \omega_l\;E_{l}(x)\;, 
\end{equation}
where $E_{p}(x)$ is the pairwise reflectance term that forces pixels that are nearby in position, chromaticity and intensity, to be assigned to the same surface reflectance, $E_{s}(x)$ enforces the shading smoothness, $E_{l}(x)$ penalizes extreme values of shading for too many pixels, and $x$ denotes the optimal labeling (chromaticity value) for a pixel.

\section{Experiments}
We show the potential merits of using color ratios by conducting the following experiments: (1) an ablation study of the cross color ratios on the Color Retinex performance, (2) the effect on the cluster initialization scheme, (3) the influence of the pairwise term, and (4) the performance on real world shadow removal. Experiments are provided on the real world object-level MIT intrinsics dataset~\cite{Grosse2009} and real world scene-level Intrinsic Images in the Wild (IIW) of complex indoor scenes for reflectance predictions~\cite{Bell2014}. We report the mean, median and trimean values to measure the central tendency of the error metrics to evaluate the robustness gains. Following the common practice, the local mean-squared error (LMSE) with a window size of 20 for the MIT dataset and the weighted human disagreement rate (WHDR) for IIW, as provided by the authors, are reported as error metrics. Lower values are better for both cases. Finally, we demonstrate the merits of our approach for the shadow removal application on the Image Shadow Triplets Dataset (ISTD)~\cite{Wang2018} and the Shadow Removal Dataset (SRD)~\cite{Qu2017} of real world outdoor scenes with a variety of qualitative evaluations.

\subsection{Combination with Color Retinex}
\label{sec:retinex_combination}
As an ablation study, we integrate cross color ratios, Equation~\eqref{eq:M}, into the Color Retinex algorithm. Color Retinex algorithm identifies reflectance changes by two threshold values, one for strong brightness changes and the other for strong chromaticity changes. If a pixel satisfies both thresholds, then it is classified as a reflectance change. Then, using a surface reconstruction algorithm, those pixels are re-integrated to achieve the final reflectance estimation. Although powerful, the algorithm is based on the Mondrian world hypothesis assuming piece-wise distinctly colored patches for the gradient separation. However, that is not always the case for real world images as they might contain weak color transitions or very strong photometric effects (e.g. shadow casts) that also cause strong gradient changes. Therefore, to show the merits, we integrate the ratios into the Color Retinex paradigm to provide complementary making the algorithm more robust. For the experiments, we use the Color Retinex implementation of~\cite{Grosse2009}.

For the integration, we first compute the cross color ratios by Equation~\eqref{eq:M} in $RGB$ space. Before computation, Gaussian smoothing is applied to the image. To merge the ratios and achieve a unified map, we combine the three ratios by their geometric mean: $\sqrt[\leftroot{-2}\uproot{2}{3}]{M1 \times M2 \times M3}$. It is applied in log space to avoid numerical instabilities. Note that the arithmetic mean can also be preferred in $RGB$ space. This way all color information is properly utilized. One might also use a single color ratio such as $M1$ (red-green), yet given an image dominated by blue color, the ratio will not be sufficiently responsive. Finally, a small threshold value of 0.02 is set to eliminate possible sensor artifacts and noise.

\begin{table}[h]
   \centering
   \begin{tabular}{ |c|c|c|}\hline
      & Color Retinex & CCR-assisted Color Retinex \\ \hline  \hline
     Mean (LMSE) & 0.0315 & \textbf{0.0313}  \\ \hline
     Median (LMSE) & 0.0303 & \textbf{0.0295} \\ \hline
     Trimean (LMSE) & 0.0314 & \textbf{0.0309} \\ \hline \hline  \hline
     Mean (WHDR) & 33.88 & \textbf{33.86}  \\ \hline
     Median (WHDR) & 33.10 & \textbf{33.00} \\ \hline
     Trimean (WHDR) & 33.26 & \textbf{33.14} \\ \hline
   \end{tabular}
   \caption {Combination of Color Retinex and the cross color ratios (CCR). The ratios further improve Color Retinex on all metrics and make it more robust.} 
   \label{tab:retinex_combination}
 \end{table}

The combination is achieved such that the response maps of Color Retinex and the cross color ratios are fused using the \emph{logical or} operation. Therefore, they become complementary to each other. If a pixel is labeled not reflectance by both, then it is discarded, whereas if a pixel is labeled as reflectance at least by one of the algorithms, it is marked as true reflectance change. We demonstrate our contribution on the MIT intrinsics dataset of real world objects and IIW real world indoor scenes for the reflectance predictions in Table~\ref{tab:retinex_combination}.

The results show that the cross color ratios further improve Color Retinex algorithm and performance gains are observed. All three metrics, mean, median and trimean, are improved. The improvements are more significant for median and trimean metrics such that Color Retinex becomes more robust. As a result, simply using an elementary setup, we manage to improve the reconstruction quality of the Color Retinex's reflectance estimations for both object and scene level cases. Note that the computation of the ratios do not add any overhead, is completely learning-free and realized in real time. They are computed directly from the input $RGB$ image (unsupervised). The results are particularly important as a large number of work consider Color Retinex approach in their optimizations~\cite{Barron2015,Ding2017,Li2014,Sheng2020,Bi2015,Cheng2019}. The results suggest that the future work intending to use the Color Retinex gradient constraint should also consider the cross color ratios for improved and more robust estimations.

\subsection{Adaptive Cluster Initialization}
\label{sec:cluster}
The optimization of Equation~\eqref{eq:iiw} is initialized by k-means clustering of pixel features. $RGB$ pixels are transformed into pixel intensity, red chromaticity, and green chromaticity as features, because chromaticity information perfectly separate reflectances under ideal conditions. Then, each pixel is labeled with a reflectance by the cluster centers. That imposes the global reflectance sparsity constraint~\cite{Bell2014,Gehler2011}. 

\begin{table}[h]
   \centering
   \scalebox{0.9}{
   \begin{tabular}{ |c|c|c|c|c|c|}\hline
      & Mean (LMSE) & Median (LMSE) & Trimean (LMSE)  \\ \hline
     Fixed $k = 10$ & 0.0484 & 0.0400 & 0.0438 \\ \hline
     Fixed $k = 20$ (default) & 0.0597 & 0.0434 & 0.0506 \\ \hline
     Fixed $k = 30$ & 0.0499 & 0.0397 & 0.0437 \\ \hline \hline
     Adaptive $k$ by ratios & 0.0443 &  0.0356 & 0.0388  \\ \hline
     Adaptive $k$ by ratios w/ ratio features & \textbf{0.0431} &  \textbf{0.0353} & \textbf{0.0366}  \\ \hline
   \end{tabular}}
      \caption {The effect of $k$ for the k-means algorithm for the MIT dataset. Adaptive setting of the $k$ value significantly improves the reflectance estimations.} 
         \label{tab:clustering}
 \end{table}

\begin{figure}[h]
 \centering
\includegraphics[width=0.8\linewidth]{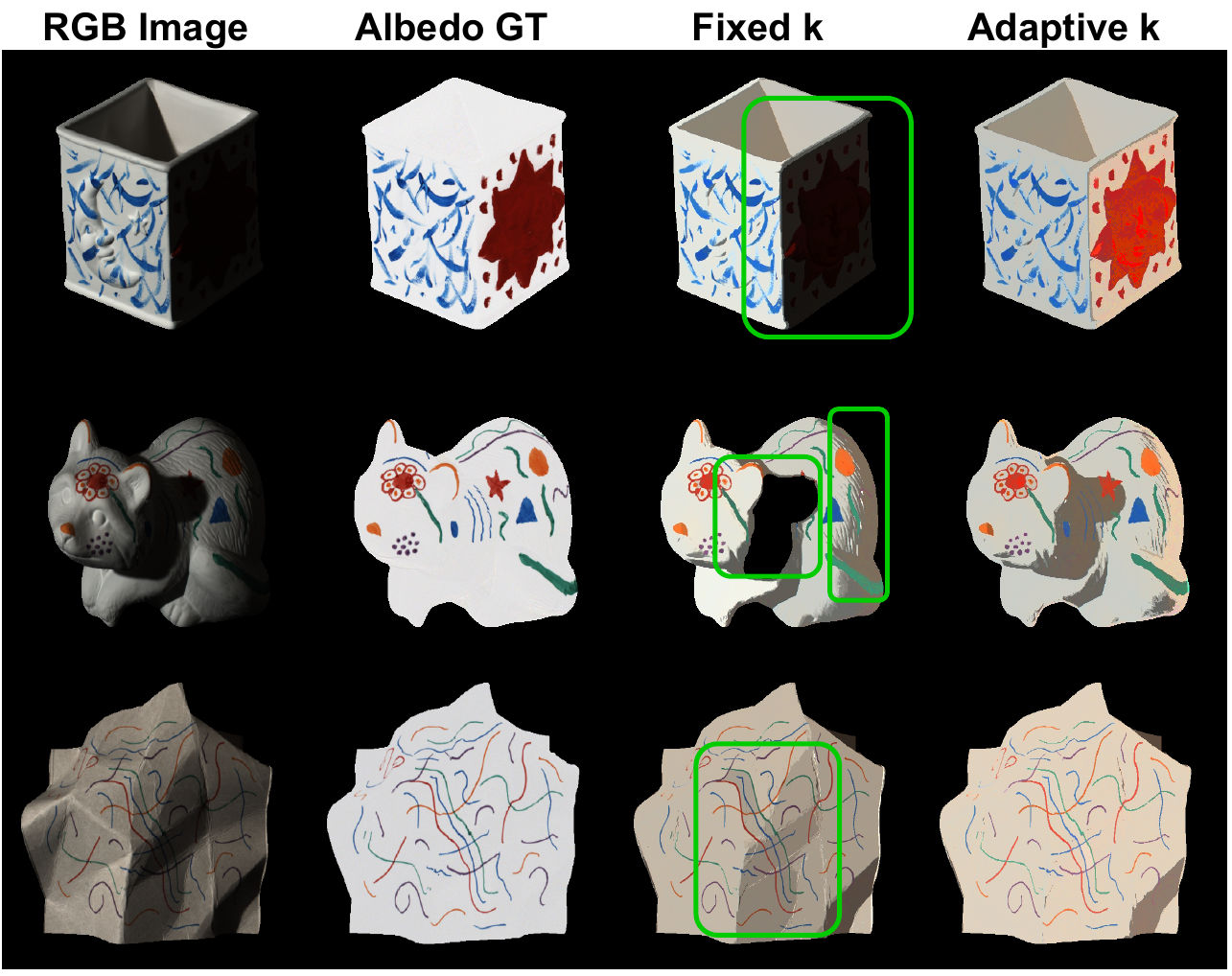}
   \caption{The effect of the ratio driven clustering. Default model extra clusters shadows and strong shadings as reflectance. Adaptive setting $k$ by the ratios makes the model more accurate and robust to photometric effects such as strong shading and shadows.}
\label{fig:clustering}
\end{figure}

Mostly, the number of clusters is treated as a hyper-parameter and decided through a grid search. However, the process is costly, not effective and dataset dependent. It needs to be adjusted per domain. The task is an active area of research (e.g.~\cite{Yan2005,Li2020}). Here, we take a different approach and use the cross color ratios to adaptively determine the number of clusters. In a homogeneously (single) colored surface, the ratios are constant (i.e. 1). They significantly deviate from 1 when there is a color change. Thus, the number of distinct color ratio values correspond to the number of distinct colors in an image. Therefore, the number of distinct color ratio values can be used to determine the number of (true color) clusters per image. To take advantage of this property, we represent each $RGB$ pixel $x$ as its cross color ratios $(M1(x)\;,\;M2(x)\;,\;M3(x))$ computed by Equation~\eqref{eq:M}. When at least one of the three features significantly deviates from 1, we count it as a unique color. Same ratios are only counted once. We define the significance by the \emph{round} function. As a result, the number of clusters can be adaptively set for each image, unlike other methods that use the same (fixed) number of clusters for all images. Finally, we include cross color ratios as features into the clustering. The ratios are first normalized by the maximum value, then weighted by 0.5 for MIT, 10 for IIW experiments. Our contribution is presented in Table~\ref{tab:clustering} and demonstrated in Figure~\ref{fig:clustering} for the MIT dataset and in Table~\ref{tab:clustering_iiw} for IIW. An advanced Retinex model~\cite{Xu2020} is also included as a comparison.

  \begin{table}[h]
   \centering
   \scalebox{0.9}{
   \begin{tabular}{ |c|c|c|c|c|c|}\hline
      & Mean (WHDR) & Median (WHDR) & Trimean (WHDR)  \\ \hline
     STAR~\cite{Xu2020} & 31.88 & 32.24 & 32.05 \\ \hline \hline
     Fixed $k = 10$ & 21.23 & 18.59 & 19.58 \\ \hline
     Fixed $k = 20$ (default) & 20.08 & 17.91 & 18.26 \\ \hline
     Fixed $k = 70$ & 19.80 & 17.52 & 18.50 \\ \hline \hline
     Adaptive $k$ by ratios w/ ratio features & 19.50 &  17.06 & 17.55  \\ \hline
     Adaptive $k$ by ratios w/ ratio features$^{*}$ & \textbf{17.51} &  \textbf{15.33} & \textbf{15.87}\\
     \hline
   \end{tabular}}
   \caption {The effect of $k$ for the k-means algorithm for the IIW dataset. Adaptive setting of the $k$ value further improves the reflectance estimations for indoor scene level images. Further improvements are also achieved with a guided filter post-processing (*)~\cite{Nestmeyer2017}.} 
     \label{tab:clustering_iiw}
 \end{table}
 
For the MIT dataset, Figure~\ref{fig:clustering} shows that the default approach produces extra clusters for shadows and strong shadings. On the other hand, strong shadings are decently handled and the region under the strong shadow is visible now in the $raccoon$ object due to the adaptive setting of the $k$ value. In addition, Table~\ref{tab:clustering} shows that the ratio driven adaptive $k$ leads to more robust and accurate results improving all three metrics. Finally, additionally including the cross color ratios as features into the clustering scheme further boosts the reconstruction performance. For the IIW dataset, Table~\ref{tab:clustering_iiw} demonstrates that the ratio driven adaptive $k$ leads to more robust and accurate results also for scene level images. The improvements are more significant for median and trimean metrics such that the ratios make the algorithm more robust. Further improvements can also be achieved with a guided filter post-processing~\cite{Nestmeyer2017}.
 
Calculation of the features $(M1(x)\;,\;M2(x)\;,\;M3(x))$ is realized in real time and does not introduce any overhead. The experiments suggest that a global reflectance sparsity prior is beneficial considering the ratio driven adaptive setting of the number of clusters. Therefore, future work intending to use the global reflectance sparsity prior should also consider the adaptive setting of the number of clusters by the ratios for improved and more robust estimations.
 
\subsection{Pairwise Reflectance}
\begin{figure}[!h]
 \centering
\includegraphics[width=0.7\linewidth]{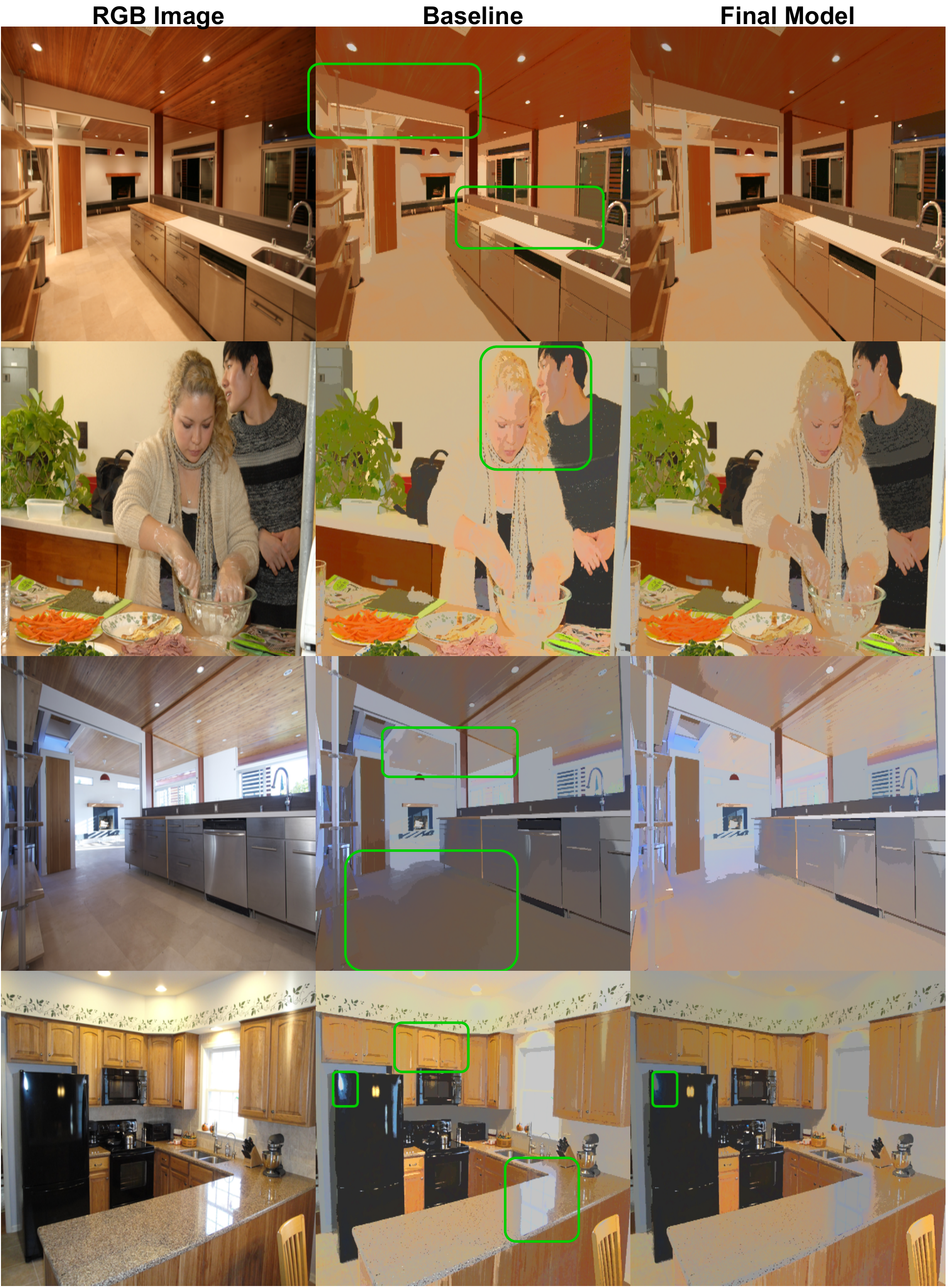}
   \caption{Reflectance evaluations for IIW dataset. Problematic parts are marked with green bounding boxes. The final model further handles discontinuities in the reflectance. It becomes more robust to direct light effects and also to specular highlights.}
\label{fig:pairwise_iiw}
\end{figure}

Finally, we inject the color ratios into the pairwise reflectance term $E_{p}(x)$ of Equation~\eqref{eq:energy_function}. The term forces pixels that are nearby in position, chromaticity and intensity to have the same reflectance value. Under white illumination, two pixels having the same reflectance should also have the same chromaticity. However, the white illumination assumption is not realistic for real world scenes. Thus, it is limited by real world applications. Therefore, we modify the function such that it also considers cross color ratios along with nearby pixels in position, pixel chromaticity and intensity. Thus, the term is further constrained by physics-based descriptors that are not limited in any assumption such as the white light. The ratio feature is defined as the geometric mean of the three cross color ratios, same as in Section~\ref{sec:retinex_combination}. We further weight it with a Gaussian to be suitable for CRF. The results are provided in Figure~\ref{fig:pairwise_iiw} for the scene level IIW estimations. It can be observed that the ratios further handles discontinuities in the reflectance map estimations making them more robust to direct light effects and also to specular highlights.

\begin{table}[h]
   \centering
   \begin{tabular}{ |c|c|c|c|c|c|}\hline
      & Mean (LMSE) & Median (LMSE) & Trimean (LMSE)  \\ \hline
     STAR~\cite{Xu2020} & 0.0478 & 0.0438 & 0.0451 \\ \hline \hline
     Default model & 0.0597 & 0.0434 & 0.0506 \\ \hline
     Final Model & \textbf{0.0424} & \textbf{0.0325} & \textbf{0.0349}  \\ \hline
   \end{tabular}
   \caption {The effect of the additional pairwise term for the MIT dataset. It further improves the reflectance estimations on all metrics and make the model more robust.} 
   \label{tab:pairwise}
 \end{table}

\begin{figure}[h]
 \centering
\includegraphics[width=1\linewidth]{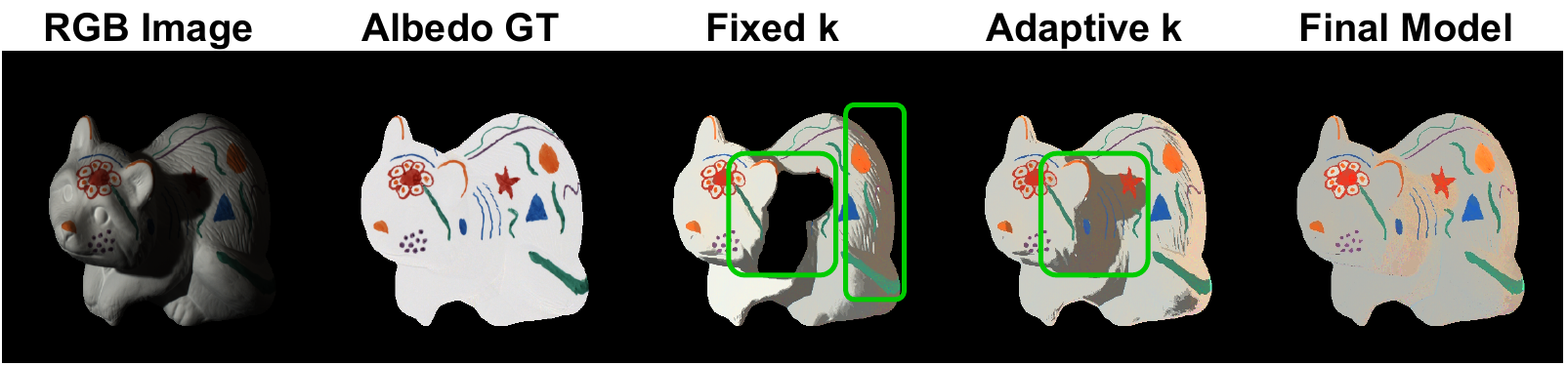}
   \caption{The additional effect of the cross color ratios as a pairwise term. The full model has significant degree of photometric invariance, capable of handling strong shadows.}
\label{fig:pairwise_mit}
\end{figure}

In addition, our final model contribution employing also adaptive k-means is presented in Figure~\ref{fig:pairwise_mit} and in Table~\ref{tab:pairwise} for the MIT dataset. An advanced Retinex model~\cite{Xu2020} is included as a comparison. The quantitative results demonstrate that the final model achieves significantly better results on all metrics thanks to the cross color ratio injections. The visual results demonstrate a significant degree of photometric invariance, capable of even handling strong shadow casts. 
 
\subsection{Scene Level Real World Shadow Removal Application}
\begin{figure}[h]
 \centering
\includegraphics[width=1\linewidth]{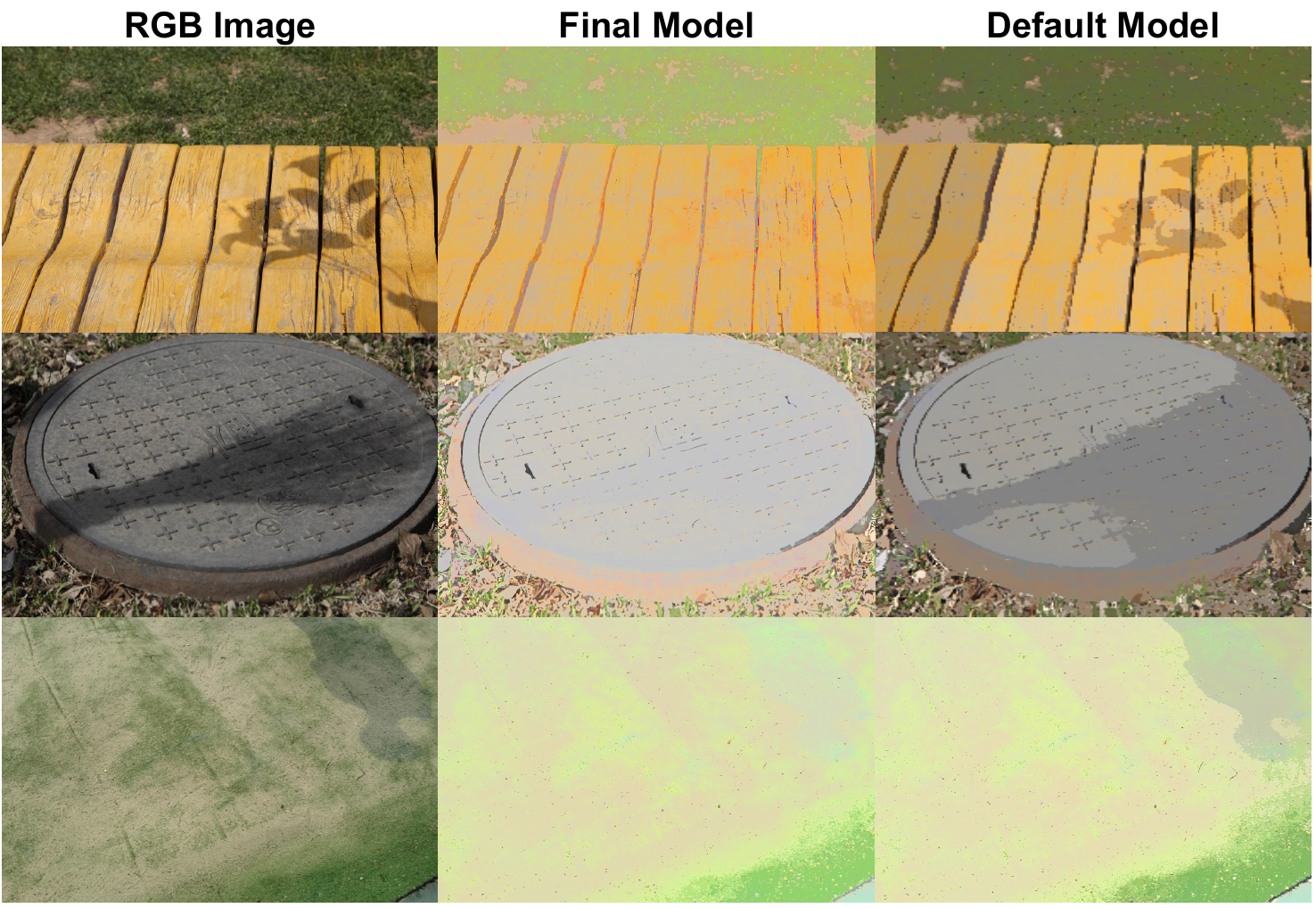}
   \caption{The default model completely fails handling shadow casts. The final proposed model driven by the photometric invariant color ratios is more robust to natural outdoor real world shadow handling. It can now differentiate drastic changes in pixel values and attribute them to the related intrinsics.}
\label{fig:shadow}
\end{figure}

In this section, we evaluate the proposed algorithm on a real world application. Reflectance images being photometric invariants should be shadow free by definition. Therefore, we demonstrate the significant benefits that the ratios bring to the reflectance estimations' shadow handling performance. To this end, we provide qualitative results on the Image Shadow Triplets Dataset (ISTD)~\cite{Wang2018} and the Shadow Removal Dataset (SRD)~\cite{Qu2017}. The results are presented in Figure~\ref{fig:shadow}. It can be observed that the photometric invariant ratios further robustify the model. The final model can handle strong shadow casts in real world natural scenes significantly better. We can even handle shadows on monochromatic surfaces as illustrated in the second row. The default model fails and produces strong shadow artefacts polluting the reflection estimations in all cases.

\subsection{Comparison with Prior State-of-the-Art}
\begin{figure}[!h]
 \centering
\includegraphics[width=1\linewidth]{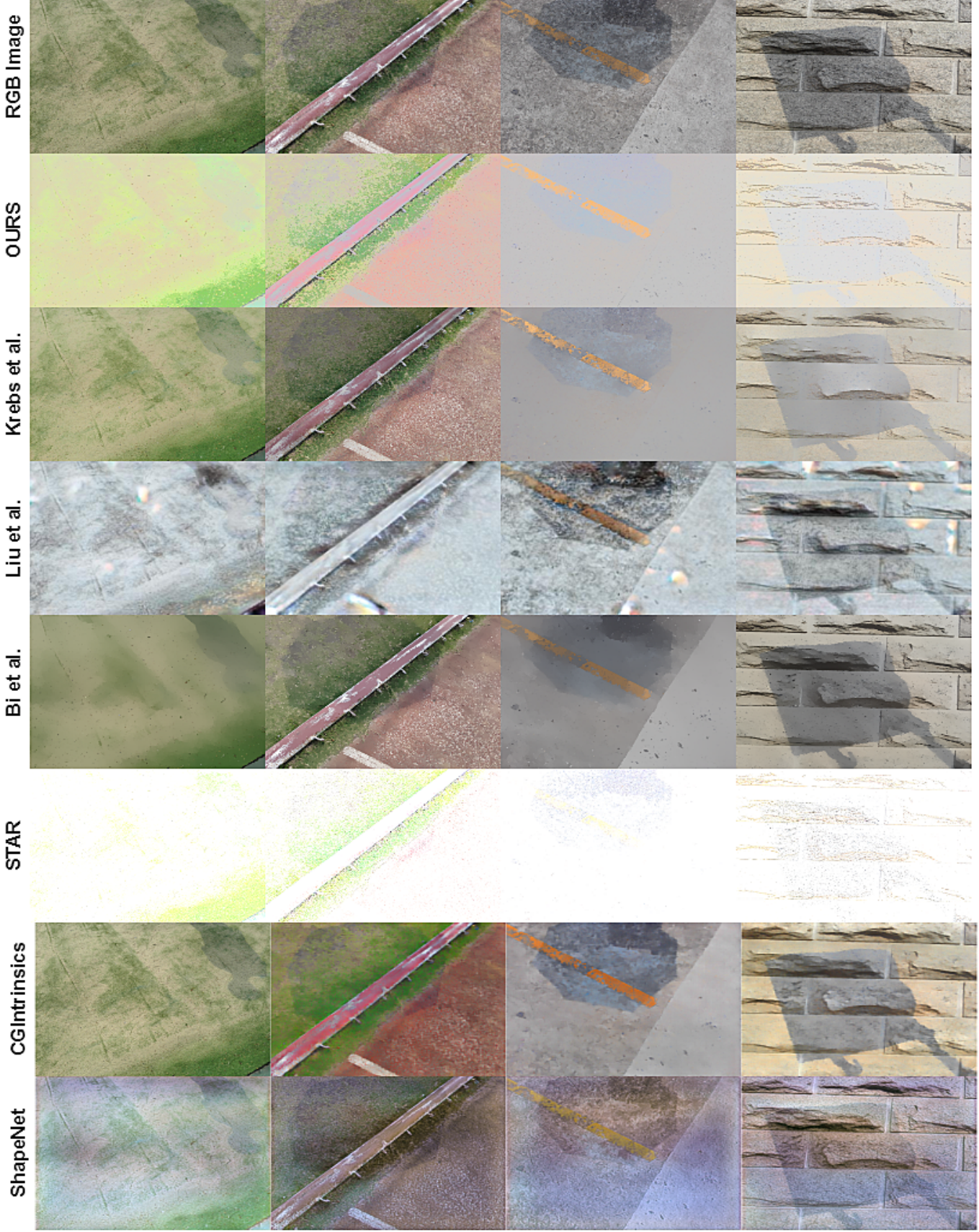}
   \caption{State-of-the-art comparisons on shadow cast handling on ISTD~\cite{Wang2018}. Both unsupervised and supervised models fail to handle shadows. Our method driven by the photometric invariant color ratios is more robust containing significantly less shadow cues and less shadow artefacts, the colors are more vivid and realistic, and the structures are well-preserved. Images are best viewed in color and on the electronic version.}
\label{fig:sota_ISTD}
\end{figure}

\begin{figure}[!h]
 \centering
\includegraphics[width=1\linewidth]{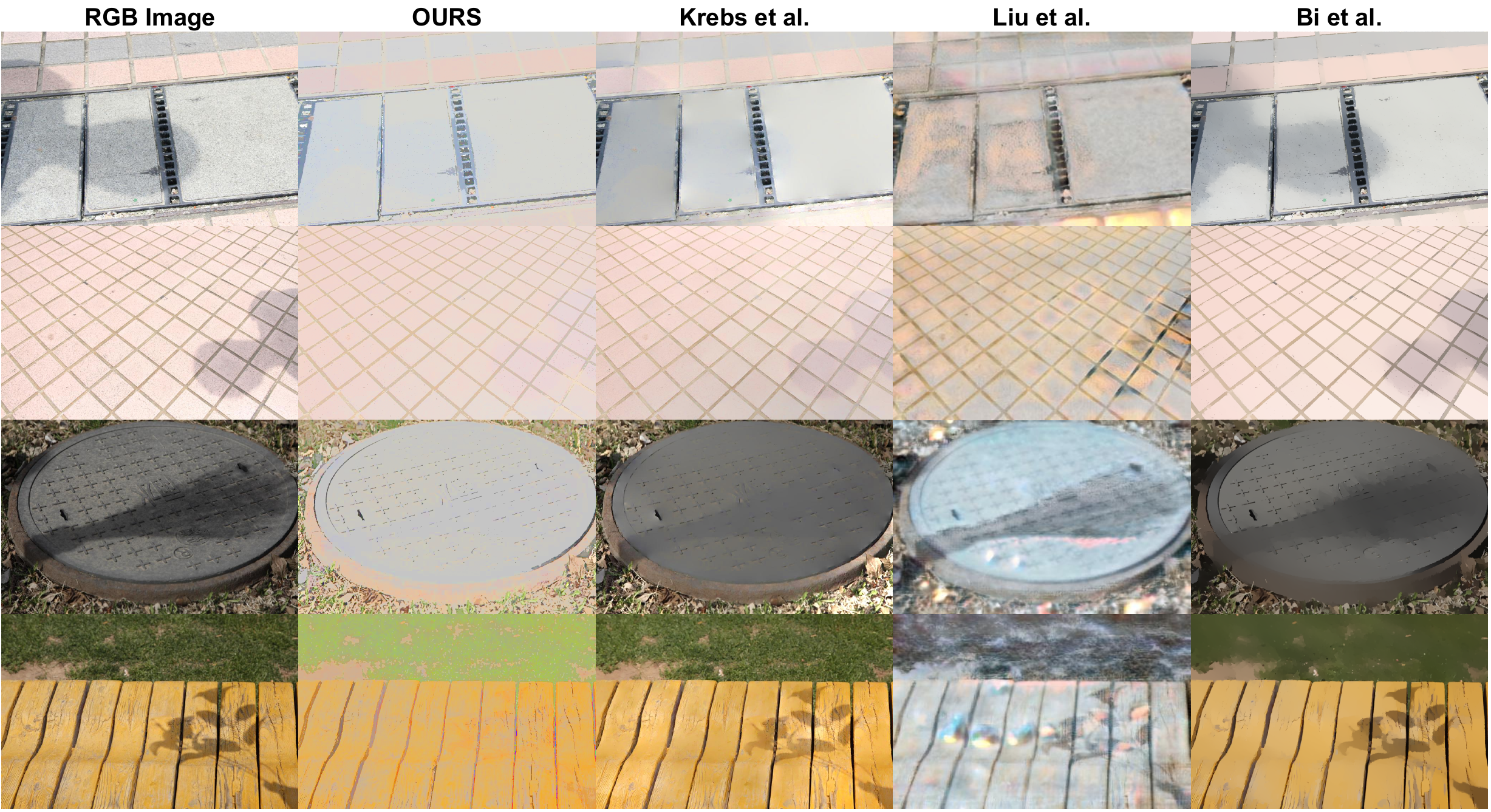}
   \caption{State-of-the-art comparisons on shadow cast handling on SRD~\cite{Qu2017}. Compared with others, our albedo estimations contain very little or almost no shadow artefacts, the colors are more vivid and realistic, and the structures are well-preserved. Images are best viewed in color and on the electronic version.}
\label{fig:sota_SRD}
\end{figure}

Finally, in this section, we provide deep qualitative comparison on ISTD and SRD with the state-of-the-art methods: (1) Krebs~\etal~recover the parameters of the dichromatic reflection model using quadratic programming~\cite{Krebs2020}, (2) Liu~\etal~introduce an unsupervised intrinsic image decomposition model that learns the latent features of reflectance and shading from unsupervised and uncorrelated data~\cite{Liu2020}, (3) Bi~\etal~use edge-preserving smoothing, Dirichlet Process Gaussian Mixture Mode for adaptive clustering, and a superpixel based CRF optimization~\cite{Bi2015}, (4) STAR is an advanced structure and texture aware Retinex model~\cite{Xu2020}. Further, we provide two supervised deep learning models: (5) CGIntrinsics combines 4 different scene-level real and synthetic datasets for training~\cite{Li2018ECCV}, (6) ShapeNet boosts the intrinsic correlations with special decoder links and is trained on 2.5M synthetic images~\cite{Shi2017}. Thus, the evaluation is profoundly diverse. The results are provided in Figure~\ref{fig:sota_ISTD} for ISTD and Figure~\ref{fig:sota_SRD} for SRD.
\newpage

The evaluations further prove the quality of the proposed invariant descriptors for natural outdoor shadow handling. Figure~\ref{fig:sota_ISTD} shows that Krebs~\etal~tend to produce smooth reflectance maps, but cannot handle shadow casts. Liu~\etal~fail to generate proper colors. The images appear way too dull and in the first two columns the scenes have lost its intrinsic color. It also tends to generate bright yellowish artefacts. Bi~\etal~using CRF optimization and adaptive clustering appear to be the closest work to ours. Yet, their method also fails to handle shadow casts. Their adaptive clustering even generates additional shadow-like artefacts on the first image. Thus, the contribution of the color ratios are more clear that our model does not just choose the cluster number adaptively, but the descriptors further constrain the reflectance predictions. STAR generates too bright images that most of the structures and colors are not visible anymore. Similarly, supervised deep learning models CGIntrinsics and ShapeNet fail to generate shadow free albedo images. ShapeNet produces further color artefacts and distortions. 

Similar behaviour is also observed on Figure~\ref{fig:sota_SRD} for SRD. Krebs~\etal~tend to produce smooth reflectance maps, but fails to handle shadow casts. Liu~\etal~again fail to generate proper colors. For example, the yellow color of the bench in the last row is completely washed away. Moreover, their model tends to generate bright yellowish artefacts contaminating the reflectance maps. Finally, Bi~\etal~results are similar to Krebs~\etal~producing smooth reflectance maps, but they cannot properly handle shadow casts.

On the other hand, our method driven by the photometric invariant color ratios, is more robust in handling natural outdoor real world shadow casts for two different domains of ISTD and SRD. Compared with others, our reflectance predictions contain significantly less shadow cues and less shadow artefacts. Furthermore, the colors are more vivid and realistic, and the structures are well-preserved and not polluted with undesired color artefacts. The results provided on two distinct dataset domains further demonstrate the generalization capability of the photometric invariant color ratios. In addition, the results relieve a surprising fact that none of the state-of-the-art models is able to handle natural outdoor shadow casts. 

\section{Conclusion and Future Work}
We demonstrated the concept of physics-based invariant descriptors used in a CRF optimization framework to improve reflectance predictions by making them more accurate and more robust to shading variations. We provided three particular ways to combine the ratios into the optimization and a final combination is presented. Experiments were provided on four datasets; a real world object level, a real world indoor scene level and two different outdoor scene level image sets with various shadow casts. Quantitative improvements were achieved on both object level and scene level indoor scenes. On the other hand, significance of the proposed method is better observed by the qualitative results. The visual results showed that the default model has no sense of shadows as it cannot differentiate if the changes in pixel intensities are caused by a color change or due to photometric effects. Incorporating the color ratios further enhances the model making it aware of the photometric cues including strong shadings and shadow casts. Thus, the model can now differentiate drastic changes in pixel values and attribute them to the related intrinsics.

For future work considering the (1) Color Retinex gradient constraint, (2) global reflectance sparsity prior, (3) pair-wise constant reflectance prior, or (4) natural shadow cast handling should also consider incorporating the photometric invariant descriptors to make their frameworks more accurate and robust. Unlike hand-crafted priors, which are based on observations and mostly computationally intensive, the color ratios are physics-based, calculated directly from the $RGB$ images in an unsupervised manner and realized in real time. The process does not involve any specialized filters or labeled data and it does not add any additional complexity to the system. Thus, further improvements are attainable without any overhead. Our future work will focus on steering deep learning models by the photometric invariant descriptors.

\section*{Acknowledgments}
This project was funded by the EU Horizon 2020 program No. 688007 (TrimBot2020). We thank Partha Das for his contribution to the experiments.

\section*{Disclosures}
``The authors declare no conflicts of interest.''


\bibliography{sample}

\begin{thebibliography}{10}
\newcommand{\enquote}[1]{``#1''}

\bibitem{Barrow1978}
H.~G. Barrow and J.~M. Tenenbaum, \enquote{Recovering intrinsic scene
  characteristics from images,} {\protect\JournalTitle{Computer Vision
  Systems}} pp. 3--26 (1978).

\bibitem{Baslamisli2018ECCV}
A.~S. Baslamisli, T.~T. Groenestege, P.~Das, H.~A. Le, S.~Karaoglu, and
  T.~Gevers, \enquote{Joint learning of intrinsic images and semantic
  segmentation,} in \emph{European Conference on Computer Vision,}  (2018).

\bibitem{Xu2019}
C.~Xu, Y.~Han, G.~Baciu, and M.~Li, \enquote{Fabric image recolorization based
  on intrinsic image decomposition,} {\protect\JournalTitle{Textile Research
  Journal}} p. 3617–3631 (2019).

\bibitem{Beigpour2011}
S.~Beigpour and J.~van~de Weijer, \enquote{Object recoloring based on intrinsic
  image estimation,} in \emph{IEEE International Conference on Computer
  Vision,}  (2011).

\bibitem{Land1971}
E.~H. Land and J.~J. McCann, \enquote{Lightness and retinex theory,}
  {\protect\JournalTitle{Journal of Optical Society of America}} pp. 1--11
  (1971).

\bibitem{Bell2014}
S.~Bell, K.~Bala, and N.~Snavely, \enquote{Intrinsic images in the wild,}
  {\protect\JournalTitle{ACM Transactions on Graphics}}  (2014).

\bibitem{Cheng2019}
Z.~Cheng, Y.~Zheng, S.~You, and I.~Sato, \enquote{Non-local intrinsic
  decomposition with near-infrared priors,} in \emph{IEEE International
  Conference on Computer Vision,}  (2019).

\bibitem{Narihia2015}
T.~Narihira, M.~Maire, and S.~X. Yu, \enquote{Direct intrinsics: Learning
  albedo-shading decomposition by convolutional regression,} in \emph{IEEE
  International Conference on Computer Vision,}  (2015).

\bibitem{Baslamisli2018CVPR}
A.~S. Baslamisli, H.~A. Le, and T.~Gevers, \enquote{Cnn based learning using
  reflection and retinex models for intrinsic image decomposition,} in
  \emph{IEEE Conference on Computer Vision and Pattern Recognition,}  (2018).

\bibitem{Gehler2011}
P.~V. Gehler, C.~Rother, M.~Kiefel, L.~Zhang, and B.~Schölkopf,
  \enquote{Recovering intrinsic images with a global sparsity prior on
  reflectance,} in \emph{Advances in Neural Information Processing Systems,}
  (2011).

\bibitem{Shen2011}
L.~Shen and C.~Yeo, \enquote{Intrinsic images decomposition using a local and
  global sparse representation of reflectance,} in \emph{IEEE Conference on
  Computer Vision and Pattern Recognition,}  (2011).

\bibitem{Barron2015}
J.~T. Barron and J.~Malik, \enquote{Shape, illumination, and reflectance from
  shading,} {\protect\JournalTitle{IEEE Transactions on Pattern Analysis and
  Machine Intelligence}} pp. 1670--1687 (2015).

\bibitem{Finlayson1992}
G.~D. Finlayson, \enquote{Colour object recognition,} Master's thesis, Simon
  Fraser University (1992).

\bibitem{Shen2008}
L.~Shen, P.~Tan, and S.~Lin, \enquote{Intrinsic image decomposition with
  non-local texture cues,} in \emph{IEEE Conference on Computer Vision and
  Pattern Recognition,}  (2008).

\bibitem{Zhao2012}
Q.~Zhao, P.~Tan, Q.~Dai, L.~Shen, E.~Wu, and S.~Lin, \enquote{A closed-form
  solution to retinex with non-local texture constraints,}
  {\protect\JournalTitle{IEEE Transactions on Pattern Analysis and Machine
  Intelligence}} pp. 1437--1444 (2012).

\bibitem{Shen2013}
J.~Shen, X.~Yang, X.~Li, and Y.~Jia, \enquote{Intrinsic image decomposition
  using optimization and user scribbles,} {\protect\JournalTitle{IEEE
  Transactions on Cybernetics}} pp. 425--436 (2013).

\bibitem{Garces2012}
E.~Garces, A.~Munoz, J.~Lopez-Moreno, and D.~Gutierrez, \enquote{Intrinsic
  images by clustering,} {\protect\JournalTitle{Computer Graphics Forum}}
  (2012).

\bibitem{Jiang2010}
X.~Jiang, A.~J. Schofield, and J.~L. Wyatt, \enquote{Correlation-based
  intrinsic image extraction from a single image,} in \emph{European Conference
  on Computer Vision,}  (2010).

\bibitem{Tappen2005}
M.~F. Tappen, W.~T. Freeman, and E.~H. Adelson, \enquote{Recovering intrinsic
  images from a single image,} {\protect\JournalTitle{IEEE Transactions on
  Pattern Analysis and Machine Intelligence}} pp. 1459--1472 (2005).

\bibitem{Ding2017}
S.~Ding, B.~Sheng, Z.~Xie, and L.~Ma, \enquote{Intrinsic image estimation using
  near-l0 sparse optimization,} {\protect\JournalTitle{The Visual Computer:
  International Journal of Computer Graphics}} pp. 355--369 (2017).

\bibitem{Li2014}
Y.~Li and M.~S. Brown, \enquote{Single image layer separation using relative
  smoothness,} in \emph{IEEE Conference on Computer Vision and Pattern
  Recognition,}  (2014).

\bibitem{Sheng2020}
B.~Sheng, P.~Li, Y.~Jin, P.~Tan, and T.~Y. Lee, \enquote{Intrinsic image
  decomposition with step and drift shading separation,}
  {\protect\JournalTitle{IEEE Transactions on Visualization and Computer
  Graphics}}  (2020).

\bibitem{Bousseau2009}
A.~Bousseau, S.~Paris, and F.~Durand, \enquote{User-assisted intrinsic images,}
  {\protect\JournalTitle{ACM Transations on Graphics}}  (2009).

\bibitem{Chen2013}
Q.~Chen and V.~Koltun, \enquote{A simple model for intrinsic image
  decomposition with depth cues,} in \emph{IEEE International Conference on
  Computer Vision,}  (2013).

\bibitem{Jeon2016}
J.~Jeon, S.~Cho, X.~Tong, and S.~Lee, \enquote{Intrinsic image decomposition
  using structure-texture separation and surface normals,} in \emph{European
  Conference on Computer Vision,}  (2016).

\bibitem{Lee2012}
K.~J. Lee, Q.~Zhao, X.~Tong, M.~Gong, S.~Izadi, S.~U. Lee, P.~Tan, and S.~Lin,
  \enquote{Estimation of intrinsic image sequences from image+depth video,} in
  \emph{European Conference on Computer Vision,}  (2012).

\bibitem{Weiss2001}
Y.~Weiss, \enquote{Deriving intrinsic images from image sequences,} in
  \emph{IEEE International Conference on Computer Vision,}  (2001).

\bibitem{Gong2019}
W.~Gong, W.~Xu, L.~Wu, X.~Xie, and Z.~Cheng, \enquote{Intrinsic image sequence
  decomposition using low-rank sparse model,} {\protect\JournalTitle{IEEE
  Access (Volume: 7)}} pp. 4024--4030 (2019).

\bibitem{Laffont2015}
P.~Y. Laffont and J.~C. Bazin, \enquote{Intrinsic decomposition of image
  sequences from local temporal variations,} in \emph{IEEE International
  Conference on Computer Vision,}  (2015).

\bibitem{Matas1995}
J.~Matas, R.~Marik, and J.~Kittler, \enquote{On representation and matching of
  multi-coloured objects,} in \emph{IEEE International Conference on Computer
  Vision,}  (1995).

\bibitem{Nayar1996}
S.~K. Nayar and R.~M. Bolle, \enquote{Reflectance based object recognition,}
  {\protect\JournalTitle{International Journal of Computer Vision}} pp.
  219--240 (1996).

\bibitem{Barnard2000}
K.~Barnard and G.~D. Finlayson, \enquote{Shadow identification using colour
  ratios,} in \emph{Color and Imaging Conference,}  (2000).

\bibitem{Gevers2001}
T.~Gevers and A.~Smeulders, \enquote{Color constant ratio gradients for image
  segmentation and similarity of texture objects,} in \emph{IEEE Conference on
  Computer Vision and Pattern Recognition,}  (2001).

\bibitem{Gevers1997}
T.~Gevers and A.~Smeulders, \enquote{Object recognition based on photometric
  color invariants,} in \emph{Scandinavian Conference on Image Analysis,}
  (1997).

\bibitem{Shi2017}
J.~Shi, Y.~Dong, H.~Su, and S.~X. Yu, \enquote{Learning non-lambertian object
  intrinsics across shapenet categories,} in \emph{IEEE Conference on Computer
  Vision and Pattern Recognition,}  (2017).

\bibitem{Li2018CVPR}
Z.~Li and N.~Snavely, \enquote{Learning intrinsic image decomposition from
  watching the world,} in \emph{IEEE Conference on Computer Vision and Pattern
  Recognition,}  (2018).

\bibitem{Lettry2018}
L.~Lettry, K.~Vanhoey, and L.~van Gool, \enquote{Darn: a deep adversarial
  residual network for intrinsic image decomposition,} in \emph{IEEE Winter
  Conference on Applications of Computer Vision,}  (2018).

\bibitem{Baslamisli2019}
A.~S. Baslamisli, P.~Das, H.~A. Le, S.~Karaoglu, and T.~Gevers,
  \enquote{Shadingnet: Image intrinsics by fine-grained shading decomposition,}
  in \emph{arXiv preprint arXiv:1912.04023,}  (2019).

\bibitem{Sial2020}
H.~A. Sial, R.~Baldrich, and M.~Vanrell, \enquote{Deep intrinsic decomposition
  trained on surreal scenes yet with realistic light effects,}
  {\protect\JournalTitle{Journal of the Optical Society of America A}}  (2020).

\bibitem{Shafer1985}
S.~Shafer, \enquote{Using color to separate reflection components,}
  {\protect\JournalTitle{Color research and applications}} pp. 210--218 (1985).

\bibitem{Finlayson1994}
G.~D. Finlayson, M.~S. Drew, and B.~V. Funt, \enquote{Color constancy:
  generalized diagonal transforms suffice,} {\protect\JournalTitle{Journal of
  the Optical Society of America A}}  (1994).

\bibitem{Grosse2009}
R.~Grosse, M.~K. Johnson, E.~H. Adelson, and W.~T. Freeman, \enquote{Ground
  truth dataset and baseline evaluations for intrinsic image algorithms,} in
  \emph{IEEE International Conference on Computer Vision,}  (2009).

\bibitem{Wang2018}
J.~Wang, X.~Li, L.~Hui, and J.~Yang, \enquote{Stacked conditional generative
  adversarial networks for jointly learning shadow detection and shadow
  removal,} in \emph{IEEE Conference on Computer Vision and Pattern
  Recognition,}  (2018).

\bibitem{Qu2017}
L.~Qu, J.~Tian, S.~He, Y.~Tang, and R.~W.~H. Lau, \enquote{Deshadownet: A
  multi-context embedding deep network for shadow removal,} in \emph{IEEE
  Conference on Computer Vision and Pattern Recognition,}  (2017).

\bibitem{Bi2015}
S.~Bi, X.~Han, and Y.~Yu, \enquote{An l1 image transform for edge-preserving
  smoothing and scene-level intrinsic decomposition,}
  {\protect\JournalTitle{ACM Transactions on Graphics}}  (2015).

\bibitem{Yan2005}
M.~Yan, \enquote{Methods of determining the number of clusters in a data set
  and a new clustering criterion,} Ph.D. thesis, Virginia Tech (2005).

\bibitem{Li2020}
X.~Li, W.~Liang, X.~Zhang, S.~Qing, and P.~C. Chang, \enquote{A cluster
  validity evaluation method for dynamically determining the near-optimal
  number of clusters,} {\protect\JournalTitle{Soft Computing}} pp. 9227--9241
  (2020).

\bibitem{Xu2020}
J.~Xu, Y.~Hou, D.~Ren, L.~Liu, F.~Zhu, M.~Yu, H.~Wang, and L.~Shao,
  \enquote{Star: A structure and texture aware retinex model,}
  {\protect\JournalTitle{IEEE Transactions on Image Processing}} pp. 5022--5037
  (2020).

\bibitem{Nestmeyer2017}
T.~Nestmeyer and P.~V. Gehler, \enquote{Reflectance adaptive filtering improves
  intrinsic image estimation,} in \emph{IEEE Conference on Computer Vision and
  Pattern Recognition,}  (2017).

\bibitem{Krebs2020}
A.~Krebs, Y.~Benezeth, and F.~Marzani, \enquote{Intrinsic rgb and multispectral
  images recovery by independent quadratic programming,}
  {\protect\JournalTitle{PeerJ Computer Science}} p. 6:e256 (2020).

\bibitem{Liu2020}
Y.~Liu, Y.~Li, S.~You, and F.~Lu, \enquote{Unsupervised learning for intrinsic
  image decomposition from a single image,} in \emph{IEEE Conference on
  Computer Vision and Pattern Recognition,}  (2020).

\bibitem{Li2018ECCV}
Z.~Li and N.~Snavely, \enquote{Cgintrinsics: Better intrinsic image
  decomposition through physically-based rendering,} in \emph{European
  Conference on Computer Vision,}  (2018).

\end{thebibliography}






\end{document}